\documentclass[letterpaper, 10 pt, conference]{ieeeconf}  

\IEEEoverridecommandlockouts                              

\title{\LARGE \bf
Learning to Imagine Manipulation Goals for Robot Task Planning
}

\author{Chris Paxton$^{1}$, Kapil Katyal$^{1}$, Christian Rupprecht$^{2}$,  Raman Arora$^{1}$,
  Gregory D. Hager$^{1}$
\thanks{$^{1}$Department of Computer Science, The Johns Hopkins University, Baltimore, MD, USA.
        Email: {\tt\small cpaxton@jhu.edu, kkatyal2@jhu.edu, hager@cs.jhu.edu, arora@cs.jhu.edu}}
\thanks{$^2$Technical University of Munich, Munich, Bavaria, Germany. Email: {\tt\small christian.rupprecht@in.tum.de}}
}

\usepackage{times}
\usepackage{helvet}
\usepackage{courier}
\usepackage{latexsym}
\usepackage[english]{babel}
\usepackage{verbatim}
\usepackage{amsmath}
\usepackage{amsfonts}
\usepackage{relsize}
\usepackage{amssymb}  
\usepackage{algorithm}
\usepackage{algpseudocode}
\usepackage{epstopdf}
\usepackage{listings}
\usepackage{longtable}
\usepackage{textcomp}
\usepackage{url}
\usepackage{alltt}
\usepackage{caption}
\usepackage{subcaption}
\usepackage{graphicx}
\usepackage{color}

\newcommand{\eat}[1]{}

\usepackage{graphicx}

\newcommand{\citep}[1]{\cite{#1}}

\begin{document}

\maketitle
\thispagestyle{empty}
\pagestyle{empty}


\begin{abstract}Prospection is an important part of how humans come up with new task plans, but has not been explored in depth in robotics.
Predicting multiple task-level is a challenging problem that involves capturing both task semantics and continuous variability over the state of the world.
Ideally, we would combine the ability of machine learning to leverage big data for learning the semantics of a task,
while using techniques from task planning to reliably generalize to new environment.
In this work, we propose a method for learning a model encoding just such a representation for task planning.
We learn a neural net that encodes the $k$ most likely outcomes from high level actions from a given world.
Our approach creates comprehensible task plans that allow us to predict changes to the environment many time steps into the future.
We demonstrate this approach via application to a stacking task in a cluttered environment,
where the robot must select between different colored blocks while avoiding obstacles, in order to perform a task. We also show results on a simple navigation task. Our algorithm generates realistic image and pose predictions at multiple points in a given task.

\end{abstract}


\IEEEpeerreviewmaketitle

\section{Introduction}

How can we allow robots to plan as humans do?
Humans are masters at solving problems. 
When attempting to solve a difficult problem, we can picture what effects our
actions will have, and what the consequences will be.
Some would say this act --- the act of prospection --- is the essence of
intelligence~\citep{prospection}.

Consider the task of stacking a series of colored blocks in a particular pattern as explored in prior work~\citep{duan2017one,xu2017neural}.
A traditional planner would view this as a sequence of high-level actions, such as \texttt{pickup(block)}, \texttt{place(block,on\_block)}, and so on. The planner will then decide which object gets picked up and in which order.
Such tasks are often described using a formal language such as the Planning Domain Description Language (PDDL)~\citep{ghallab1998pddl}.
To execute such a task on a robot, specific goal conditions and cost functions must be defined, and the preconditions and effects of the each action must be specified -- which is a large and time consuming undertaking~\citep{beetz2012cognition}.
Humans, on the other hand, do not require that all of this information be given beforehand. We can learn models of task structure purely from observation or demonstration. We work directly with high dimensional data such as images, and can reason over complex paths without being given an explicit structure.

As a result, there has been much interest in learning prospective models for planning and action.
Deep generative models such as conditional GANS~\citep{DBLP:journals/corr/IsolaZZE16} or multiple-hypothesis models for image prediction~\citep{rupprecht2017learning,chen2017photographic} allow us to generate realistic future scenes. In addition, 
a recent line of work in robotics focuses on making structured prediction~\citep{finn2016deep,finn2016unsupervised}; 
\citep{byravan2017se3} proposed SE3-nets, which predict object motion masks and six degree of freedom movement for each object; \citep{ghadirzadeh2017deep} predict trajectories to move to intermediate goals.
However, so far these approaches focus on making relatively short-term predictions, and do not take into account variability in the ways a task can be performed in a stochastic world. 

In general, deep policy learning has proven successful at learning well-scoped, short horizon robotic tasks~\citep{levine2016end,sung2016robobarista,finn2016unsupervised}.
Recent work on one-shot imitation learning learned general-purpose models for manipulating blocks, but relies on a task solution from a human expert and does not generate prospective future plans for reliable performance in new environments~\citep{duan2017one,xu2017neural}. These are very data intensive as a result:~\citep{duan2017one} used 140,000 demonstrations.

\begin{figure*}[bt]
  \centering
  \includegraphics[width=2\columnwidth]{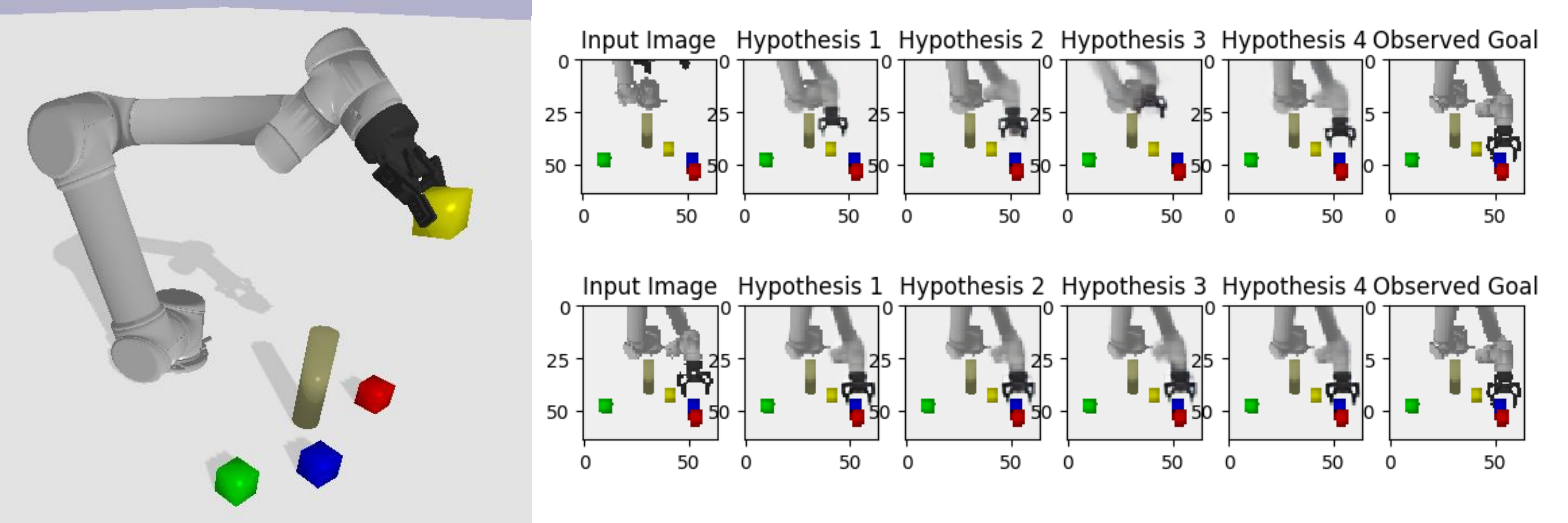}
  \caption{A simple stacking task including an obstacle that must be avoided.
  The robot must decide which blocks to pick up and move, and which block to
  put them on, taking into account its workspace and the obstacle. The right side shows how predictions change as the robot moves.}
  \label{fig:blocks-example}
\end{figure*}

Instead, we propose a model that learns this high level task structure and uses it to generate interpretable task plans by predicting sequences of movement goals.
These movement goals can then be connected via traditional trajectory optimization or motion planning approaches that can operate on depth data without a semantic understanding of the world. Fig.~\ref{fig:blocks-example} shows the task as well as predictions resulting from our algorithm at different stages.


To summarize, our contributions are:
\begin{itemize}
\item Approach for learning a predictive model over world state transitions from a large supervised dataset, suitable for task planning.
\item Analysis of the parameters that make such learning feasible in a
  stochastic world.
\item Experimental results from a simulated navigation and a block-stacking domain.
\end{itemize}

\section{Related Work}\label{sec:background}

In robotics, TAMP approaches are very effective at solving complex problems involving spatial reasoning~\citep{lagriffoul2014efficiently,toussaint2015logic}.
A subset of planners focused on Partially Observed Markov Decision Process extend this capability into uncertain worlds, such as DeSPOT~\citep{somani2013despot}.
Only a few recent works have explored integration of planning and learning.
Recent work has examined combining these approaches with QMDP-nets~\citep{karkus2017qmdp} that embed learning into a planner using a combination of a filter network and a value function approximator network in the form of a set of convolutional layers with shared weights. Similarly, value iteration networks embed a planner into a neural network which can learn navigation tasks~\citep{tamar2016value}.
\cite{chen2017neural} discuss neural network architectures for memory in
navigation.
\cite{vezhnevets2016strategic} propose the Strategic Attentive Writer (STRAW) as a sequence
prediction technique applicable to planning sequences of actions.
In~\citep{PaxtonRHK17} the authors use Monte Carlo Tree Search (MCTS) together with a set of learned action and control policies, but again do not learn a predictive
model of the future.
However, none of these approaches provide a way to predict or evaluate possible futures.

In this work, we examine the problem of learning representations for high-level
predictive models for task planning. Prediction is intrinsic to planning in a
complex world~\citep{prospection}. Robotic motion planners assume a
causal model of the world. \cite{ondruska2016deep} fit a predictive
model to predict the true state of an occluded world as it evolves
over time. \cite{lotter2016deep} propose PredNet as a way of
predicting sequences of images from sensor data, likewise with the goal being
to predict the future. \cite{finn2016unsupervised} use unsupervised learning of
visual models to push objects around in a plane.

The options framework provides a way to think of MDPs as a set of many high
level ``options,'' each active over a certain window~\citep{suttonPS99}.
Policy sketches~\citep{andreas2016modular} are one method that uses curriculum
learning with a set of ``policy sketches.''
Another option is FeUdal networks, in which a ``manager'' network sets goals
for lower level ``worker'' networks~\citep{vezhnevets2017feudal}.


\paragraph{Learning Generative Models.}\hspace*{-5pt}~Such a prediction system must be able to deal with a stochastic world.
Prior work has examined several ways of generating multiple realistic predictions~\citep{rupprecht2017learning,chen2017photographic,ghadirzadeh2017deep}.
Generative Adversarial Networks (GANs) can be used to generate samples as well.
\cite{ho2016generative} applied adversarial methods to imitation learning.
\cite{rupprecht2017learning} proposed a multiple hypothesis loss function as a way of predicting multiple possible goals when the prediction task is ambiguous
ambiguous. 
Similarly, \cite{chen2017photographic} proposed a custom loss to synthesize a diverse collection of photorealistic images.
\cite{finn2016deep} learn a deep autoencoder as a set of convolutional blocks followed by a spatial softmax; they find that this representation is useful for reinforcement learning.
More recently,~\cite{ghadirzadeh2017deep} proposed to learn a deep predictive network that uses a stochastic policy over goals for manipulation tasks, but without the goal of additionally predicting the future world state.
\cite{sung2016robobarista} learn a deep multimodal embedding for a variety of tasks; this representation allows them to adapt to appliances with different interfaces while reasoning over trajectories, natural language descriptions, and point clouds for a given task. 
Recently, \cite{higgins2017darla} propose DARLA, the DisentAngled Representation
Learning Agent, which learns useful representations for tasks that enable 
generalization to new environments.

\section{Approach}\label{sec:approach}


We define a planning problem with continuous states $x \in \mathcal{X}$ and controls $u \in \mathcal{U}$. Here, $x$ contains observed information about the world: for example, for a manipulation task, it includes the robot's end effector pose and input from a camera viewing the scene.
We augment this with high-level actions $a \in A$ that describe the task structure.
We also assume that there is some hidden world state $h$, which encodes both task information and the underlying truth of the input from the various sensors.
The symbolic world state here is not observed directly: there are numerous possible combinations of predicates that could be meaningful.

Our goal is to learn models grounding this problem as an MDP over options, so that at run time we can generate intelligent, comprehensible task plans.
Specifically, we will first learn a goal prediction function, which is a mapping $T(x,a) \rightarrow (x',a)$.
In other words, given a particular action and an observed prediction, we want to be able to predict both a continuous end goal and the actions necessary to take us there. 

We posit three components of this goal prediction function:
\begin{itemize}
	\item[1.] $f_{enc}(x,a)\rightarrow h$, a learned encoder function maps observations and descriptions to the hidden state.
	\item [2.] $f_{dec}(h) \rightarrow (x,a)$, a decoder function that maps from the hidden state of the world to the observation space.
	\item [3.] $T_{i}(h) \rightarrow h'$, the $i$-th learned world state transformation function, which maps to different positions in the space of possible hidden world states.
\end{itemize}

In addition, we assume that the hidden state $h$ consists of all the necessary information about the world. As such, we can learn two additional functions of use in planning:
\begin{itemize}
	\item $V(h)$, the expected reward-to-go from a particular hidden state
	\item $p(a | h)$, the policy over discrete high-level actions from a given hidden state
\end{itemize}

While we cannot observe the hidden world state, we can observe all of
these other fields easily, given demonstrations of a task.
Prior work has examined learning feature
encoding~\citep{parr2008analysis,song2016linear}.
Our goal is to learn a set of encoders and decoders that project our available
features into the latent world state $h$ that can be used for planning.

Our method assumes we have a dataset, containing mixed execution failures and successes, where we wish to be able to predict a large number of plausible task executions and their consequences.
We generate a large data set of action executions, with mid- and high-level
labels. We assume semantic labels have been provided for different actions to provide meaningful intermediate goals for prediction, such as \texttt{grasp(red\_block)} or
\texttt{close\_gripper}.

\subsection{Model Architecture}\label{sec:model}

\begin{figure*}[bt]
	\includegraphics[width=2\columnwidth]{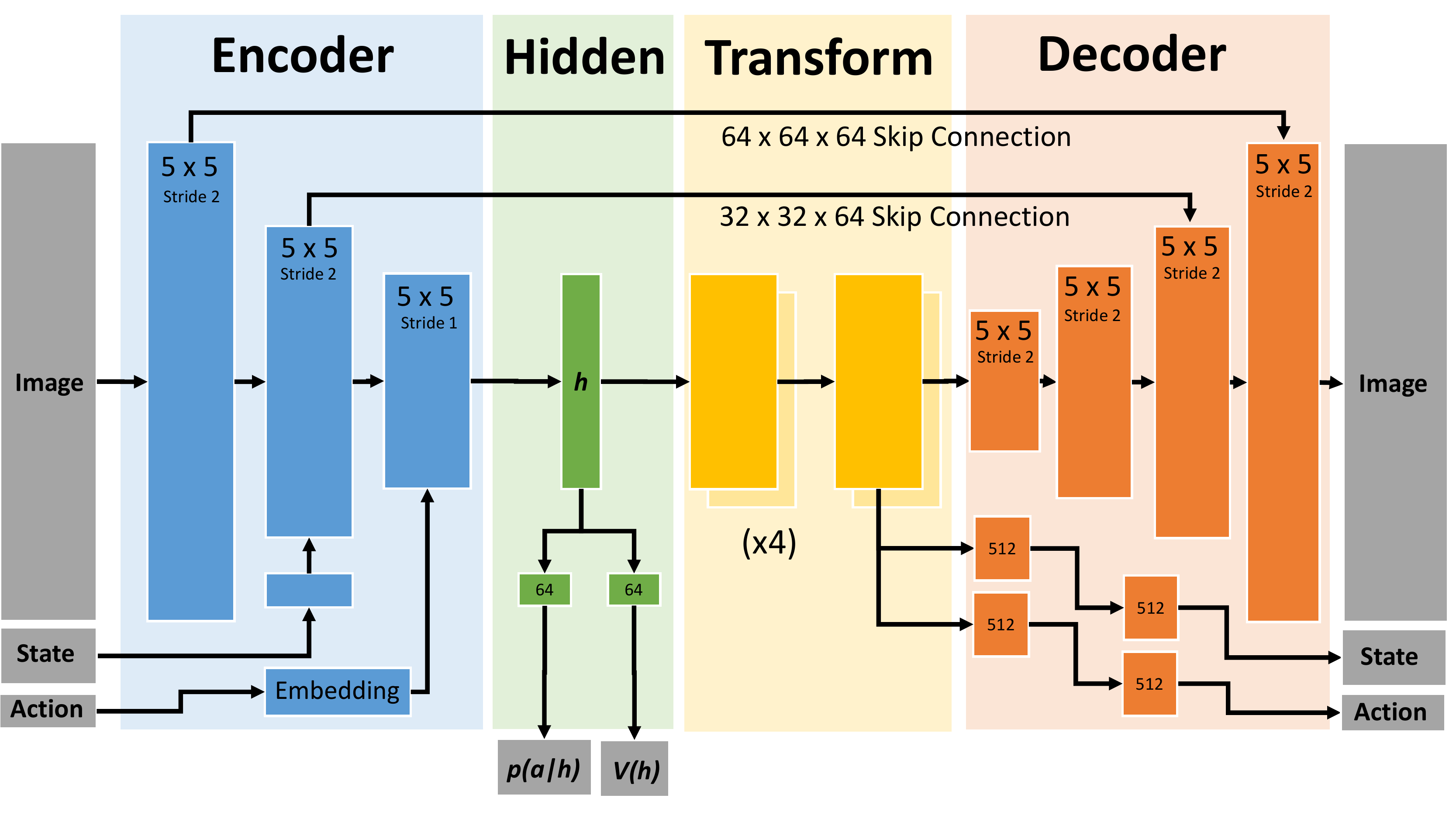}
	\caption{Overview of the predictor network. The input image and features are passed through a series of convolutional blocks in the Encoder and a spatial softmax extracts the hidden state representation. The dense layers in the Transform block compute a new world state, which is the input into the Decoder network.}
	\label{fig:predictor_net}
\end{figure*}

Fig.~\ref{fig:predictor_net} shows the architecture of our subtask goal prediction network.
The model takes in the observed high-level action $a$ and the state observations $x$.
For the manipulator example, $x = (x_I,x_{ee},x_g)$ is the image, arm
position, and gripper state, respectively; for the navigation task, $x = (x_I,x_p)$
is the image and the robot's position in the world.

The encoder $f_{enc}(x)$ is a set of $5 \times 5$ convolutional blocks. After the first convolution with stride 1, we tile input from the robot's measured state (end effector pose or robot position) as additional information. The last layers in the network are a set of spatial soft-argmax layers, as used by~\citep{finn2016deep, ghadirzadeh2017deep} and~\citep{levine2015end}. These layers extract a set of keypoints which form our ``hidden'' representation to be fed into the Transform block.
Each Transform block is composed of a set of dense layers.

We examine two alternative models for our neural network architecture: a simple encoder-decoder neural network and a U-net.
The U-net is similar to the model proposed by \cite{DBLP:journals/corr/IsolaZZE16} for image generation using conditional Generative Adversarial Networks,
and was found to generate very realistic images when performing photographic image synthesis~\citep{chen2017photographic}.

The nonlinear transformation function $T_j(h) \rightarrow h'$ maps the current hidden world representation $h$ to one of the $m$ latent representations $h'_j$ that represent possible goals the robot could pursue. We can then select between available goals to generate a task plan. This is beneficial since predicting a future state from $x$ is an  ambiguous problem; we can allow the system to predict several possible hypotheses $h'_j$ from the current state $h$. 
Since in the training data for every roll out of the task we only see one possible future instead of all of them, we need to take special care of training the system to allow it to predict multiple possibilities. 
This is achieved through the multiple hypotheses loss described in Sec.~\ref{ssec:mhp}.

Finally, in most tasks, there is some variability over specific goal poses.
In order to generate crisp, reliable hypotheses, we consider representing this variability by concatenate a vector of random noise at every step in the training process to the hidden state, before applying the transformations, with the goal being to represent the minor variability between actions. In this case, the transformation function would be expressed as $T_j(h,z)$ for a uniformly-sampled random noise vector $z$.

\subsection{Multiple Hypotheses Loss Function}\label{ssec:mhp}
Many deep learning approaches fail when learning approaches for which there are multiple, disjoint correct outputs. For this reason, recent work has explored multiple hypothesis learning for prediction~\citep{rupprecht2017learning} and for photographic image generation~\cite{chen2017photographic}.
We use a modified version of the Multiple Hypothesis (MHP) loss function \citep{rupprecht2017learning} to train our predictor model. This will predict $N_H = 4$ to $N_H = 8$ different possible future worlds.

The loss for a single hypothesis $\widehat{x}_i$ is expressed as the weighted
sum of the different outputs, where each state $x$ is expressed as a number of
different observation variables $v$, each of which is some subset of this
observed state $x$.
For example, with the manipulator robot, $v \in \left\{I, q, g\right\}$ is one
of our three classes of observations available based on the robot's current
state. For a mobile robot $x$ may consist of GPS position and camera view. 
This means that for a particular hypothesis $\widehat{x}_i$, the loss $c(\cdot,\cdot)$ is expressed as:
\[
  c(\widehat{x}_i,\widehat{a}_i) = w_a c(a) + \sum_{v \in x}w_v c(\widehat{v}_i).
\]
\noindent Here $w_v$ is the weight associated with a particular view of the world state,
and $w_a$ is the weight associated with predicting the correct low-level
action.
In practice, we use $c(x_i) = \|x_i - \widehat{x}_i\|$, the mean absolute error (MAE), when computing the loss on predicted state variables.
The MAE encourages the system to make as few mistakes as possible when predicting these errors, and has a normalizing effect that reduces small errors.
This results in sharper predictions in practice.
We handle high-level action predictions $\widehat{a}$ slightly differently.
We use an embedding for the action description to map it to a one-hot vector, and compute the cross entropy loss based on $p(a')$.

Then, we can express the loss function simply as:
\[
  c(x, a) = \min_i c\left(\widehat{x}_i,\widehat{a}_i\right).
\]

The issue with this is that in many cases some hypotheses will not make sense.
To deal with this, we consider adding an exploration probability $\lambda$,
such that with uniform probability we select one of the existing hypotheses to update.
This is equivalent to adding an average cost to the existing loss function.
Thus, the overall loss is computed as:
\[
  (1 - \lambda) \min_i c(\widehat{x}_i,\widehat{a}_i) + \dfrac{\lambda}{N_H} \sum_{i}^{N_H} c(\widehat{x}_i,\widehat{a}_i).
\]

We used $N_H = 4$ hypotheses in our experiments, with $\lambda = 0.05$.

\subsection{Value Function and Action Prior}

In addition, we learn two functions $V(h)$ and $p(a | h)$.
These are the estimated value of a given world, and the probability of taking a particular action from that world, respectively.
The value function $V(h)$ is trained using the full data set, and can be learned end-to-end with the encoder-transform-decoder architecture. The value function operates on the current hidden state $h$ and returns the expected reward-to-go -- i.e. whether or not we expect to see a possible success or failure farther on in the task if we continue down this route.

The action prior $p(a | h)$ is learned on successful training data only. The goal of this function is to tell us which actions to explore first, when performing a tree search over different possible futures. This is useful because performing a tree expansion is a relatively expensive process.
We also provide this action prior as an additional input to the transform block.




\section{Experimental Setup}\label{sec:experiment}

We applied the proposed method to both a simple navigation task using a
simulated Husky robot, and to a UR5 block-stacking task.
\footnote{Source code for all examples will be made available after publication.}

In all examples, we follow a simple process for collecting data. First, we generate a random world configuration, determining where objects will be placed in a given scene. The robot is initialized in a random pose as well. We automatically build a task model that defines a set of control laws and termination conditions, which are used to generate the robot motions in the set of training data.
Legal paths through this task model include any which meet the high-level task specification, but may still violate constraints (due to collisions or errors caused by stochastic execution).

We include both positive and negative examples in our training data.
Training was performed with Keras~\cite{chollet2015keras} and Tensorflow for 30,000 iterations on an NVidia Tesla K80 GPU, with a batch size of 32. Training took 12-18 hours, depending on the exact parameters used.


\subsection{Robot Navigation}

\begin{figure*}[bt]
  \centering
  \includegraphics[width=1.95\columnwidth]{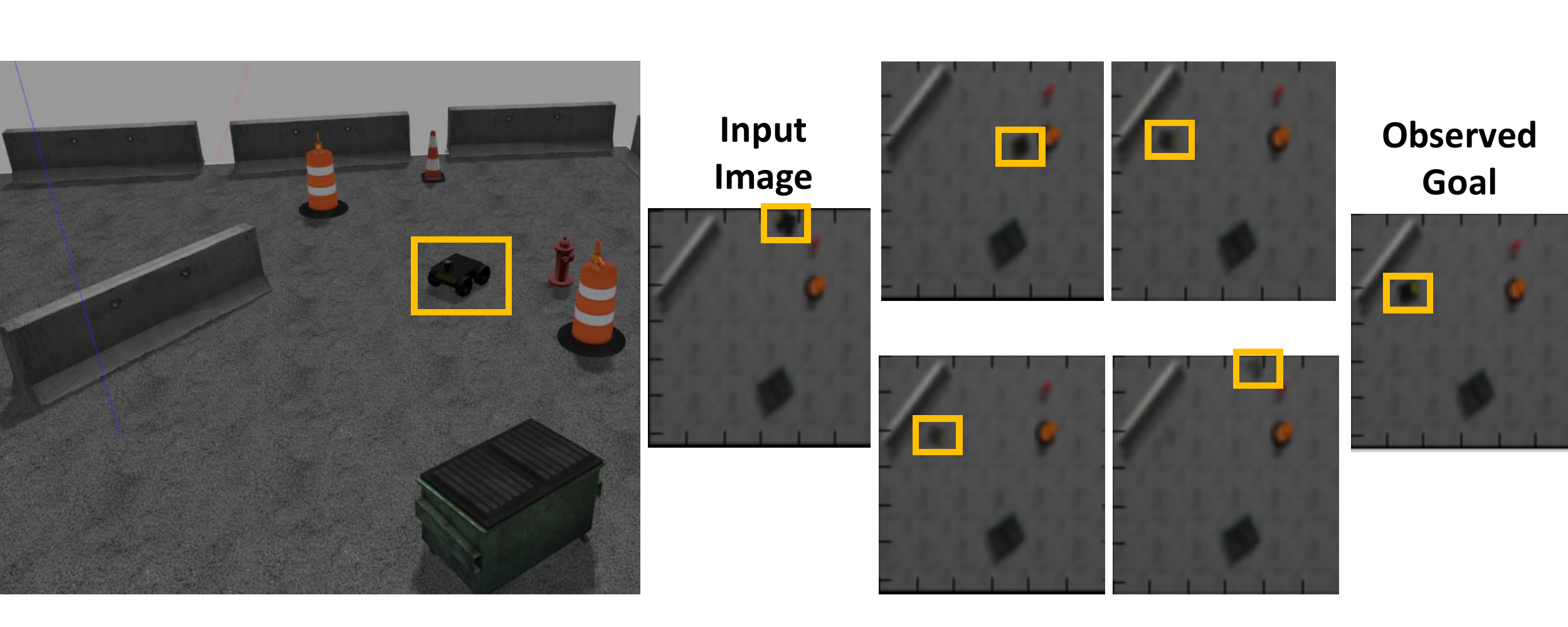}
  \caption{Prediction Results from Husky Simulation. Image on the left shows the scene in the Gazebo simulator; right images show predicted possible destinations. Robot position is highlighted.}
  \label{fig:husky_results}
\end{figure*}

In the navigation task, we modeled a Husky robot moving through a construction site environment.  The high level actions available to the robot are to investigate one of four objects: a barrel, a barricade, a constrcution pylon or a block.  
We represent the robot state by the six Degree of Freedom (6DOF) pose of the robot, represented as the concatenation of position $(x, y, z)$ and orientation $(\omega_r,\omega_p,\omega_y)$.  In addition to the robot state, we use a 64x64 RGB image taken from overhead of the scene to provide an aerial view of the environment. Data was collected using a Gazebo simulation of the robot navigating to any of the four targets.  We trained the network from Fig.~\ref{fig:predictor_net} to generate predictions of these possible goals with the number of hypotheses set to $4$.  

Fig.~\ref{fig:husky_results} shows examples of the robot in its start and (ground truth) goal pose, as well as $4$ hypothetical future predictions generated from the dataset. The robot was able to accurately predict input images to the same fidelity as the input image. We also analyzed the effect of varying network architectures on the final pose error prediction. 

\eat{
Table~\ref{table:pose_error} summarizes the final pose error averaged across the last 10 samples after training for 7500 trials.  

\begin{table*}[h]
\caption{Pose Error Analysis}
\begin{center}
 \begin{tabular}{| c | c | c | c | c |} 
 \hline
 \textbf{Label} & \textbf{Baseline} & \textbf{No Skip Connections} & \textbf{No Dropout} & \textbf{Max Dropout} \\ [0.5ex] 
 \hline\hline
 Dumpster & 0.808 & 1.997 & 1.801 & 1.945 \\
 \hline
 Barrier & 2.045 & 1.997 & 1.110 & 1.635 \\
 \hline
 Construction Barrel & 2.472 & 3.315 & 1.552 & 3.434 \\
 \hline
 Fire Hydrant & 1.859 & 3.332 & 1.780 & 1.021 \\
 \hline
\end{tabular}
\end{center}
\label{table:pose_error}
\end{table*}
}

\subsection{Block Stacking}

To analyze our ability to predict goals for task planning, we learn in a more elaborate environment.
In the block stacking task, the robot needed to pick up a colored block and place it on top of any other colored block.
To add difficulty, we place a single obstacle in the world.
The robot succeeds if it manages to stack any two blocks on top of one another and failed immediately if either it touches this obstacle or if at the end of 30 seconds the task has not been achieved. Training was performed on a relatively small number of examples: we used 7500 trials, of which 3152 were successful.

In this case, we represent the robot in terms of its 6DOF end effector pose $x_{ee}$,
encoded as the concatenation of the position $(x, y, z)$ and the roll-pitch-yaw$(\omega_r, \omega_p, \omega_y)$, such that $x_{ee} = \left\{x, y, z, \omega_r, \omega_p, \omega_y\right\}$. The state of the gripper was expressed as a single variable $x_g \in (0, 1)$. In addition, we provide a $64 \times 64$ RGB image of the scene from an external camera, referred to as $x_I$.
Fig.~\ref{fig:blocks-example} shows the results of this process.
Scenes included four blocks and the obstacle in addition to the robot, but lack any other objects or background clutter.

\eat{
Consider a minimal representation of the scene: it could be expressed in terms of $7$ robot variables ($6$ degrees of freedom for the end effector pose, plus $1$ for the state of the gripper).
In addition, there are at most $6$ degrees of freedom for each of the five objects.
This means that a conservative representation for any individual scene would be $37$ dimensions, not including change over time, contact interactions between objects, or information about the task as a whole. As such, allowing for some redundancy to handle variations in behavior and object interations, $128$ dimensions should be sufficient to capture information about the scene.
}


\section{Results}

\begin{figure}[bt]
  \centering
  \includegraphics[width=\columnwidth]{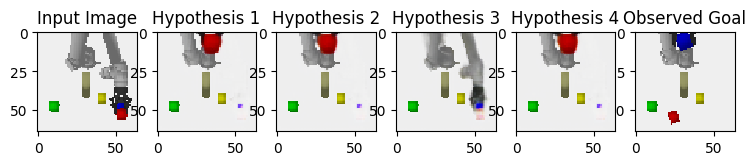}
  \caption{An example of a bad prediction. Here, the algorithm clearly attempts
           to predict what~happens when it picks up the red block, but the blue
           block prevented the gripper from properly closing.}
  \label{fig:predict-failure}
\end{figure}

Fig.~\ref{fig:blocks-example} shows examples of good predictions in at two different points in the block-stacking task: when
first choosing a block to pick up, and lifting the block over the obstacle.
Our model attempts to predict both failures and successes, but we see far
better accuracy when predicting successes on our data set due to high variability in possible failures.
Fig.~\ref{fig:predict-failure}.
shows the robot attempting to pick up red block, but the blue block interfered.
This will result in a task failure, which means the robot should attempt to pick up a
different block first instead. Such failures are random and hard to predict, in 
contrast to successes which tend to be very similar.

\begin{figure}[bt]
  \centering
  \includegraphics[width=\columnwidth]{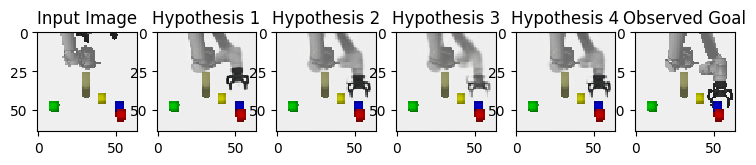}
  \includegraphics[width=\columnwidth]{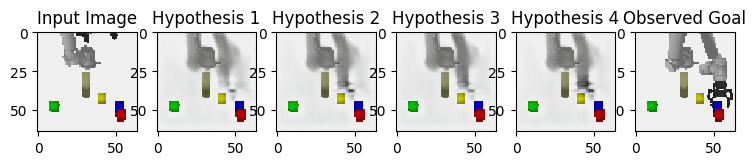}
  \caption{Predictions made using different levels of dropout to train the
  decoder network. On the top, the decoder was trained with a dropout rate of
0.125; on the bottom, the network was trained with dropout of 0.5.}
\label{fig:dropout}
\end{figure}

\begin{figure}[bt]
	\centering
	\includegraphics[width=\columnwidth]{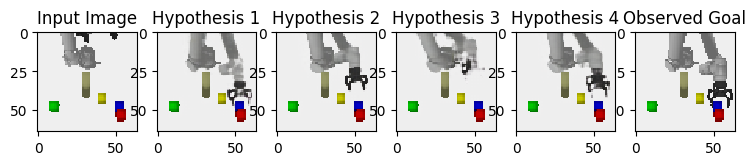}
	\includegraphics[width=\columnwidth]{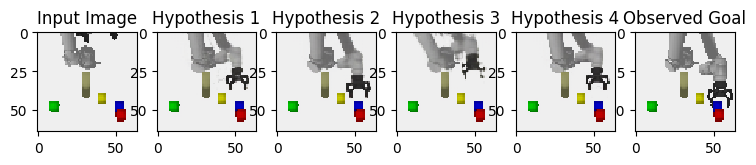}
	\caption{Concatenating a vector of random noise to the image allows
		individual transforms to better capture uncertainty in the resulting
		image. This results in crisper predicted images (bottom).}
	\label{fig:noise}
\end{figure}

\paragraph{Dropout and Stochastic Predictions.}~\hspace*{-7pt}~Differing levels of dropout have a marked effect on the quality of the models
learned during our training process. Compare the set of predictions on the top
of Fig.~\ref{fig:dropout} with those on the bottom, versus those in
Fig.~\ref{fig:noise}.

Using dropout in either the transforms or in the decoder will blur
the results, giving us a less accurate estimate of what the future world
may look like. Using insufficient dropout will likewise cause issues, because of the high accumulated variance between objects over time.

We provide a different random seed at each time step for each prediction; the
networks are able to use this noise during execution to capture some additional
level of uncertainty. A dropout level of $0.125 - 0.25$ provided the best results;
for future versions of the model, we use these dropout levels.

\paragraph{Representing Randomness.}~\hspace*{-7pt}~We represent randomness in two ways: (1) as the discrete choice of which
hypothesis occurs, and (2) as a vector of additional noise concatenated to the
hidden representation. We find that both of these are useful for representing
randomness in the resulting world state, in a way that gives us relatively crisp,
realistic images.

Fig.~\ref{fig:noise} shows the effect on models with no dropout. On the top we see
a model trained without this random vector; on the bottom we see one trained
with it. The addition of random noise allows the model to generate cleaner
predictions.
Our results show that capturing randomness may help learning predictive task models, but results are somewhat inconclusive: on training experiments with a large amount of dropout, these made little difference. With relatively low levels of dropout (e.g. dropout of $0.125$), we saw that adding a 32-D noise vector resulted in validation loss of $0.0125$ vs. $0.0150$ when including negative examples, though we did not observe the same effect with higher levels of dropout.
This sort of randomness makes a small difference under different conditions,
and led to better generalization and improved test performance.
Adding 32 random noise dimensions to our final model saw image loss of $1.21 \times 10^{-4}$ and pose error of $5.51 \times 10^{-5}$ after 300,000 training steps, while not adding this random noise had a comparable image loss ($1.85 \times 10^{-4}$) but a noticeably higher average pose error on test data ($8.60 \times 10^{-5}$).
These findings suggest that adding this vector is unimportant for ``big picture'' details, but that it helps capture small, local variance.

We compared against one additional method for representing randomness within each mode: predicting a mean and a covariance for each transform, which necessitated adding a KL loss term to the proposed loss. However, this resulted both in worse performance and resulted in the collapse of the separate modes.

\begin{figure*}[bt]
	\centering
	\includegraphics[width=2\columnwidth]{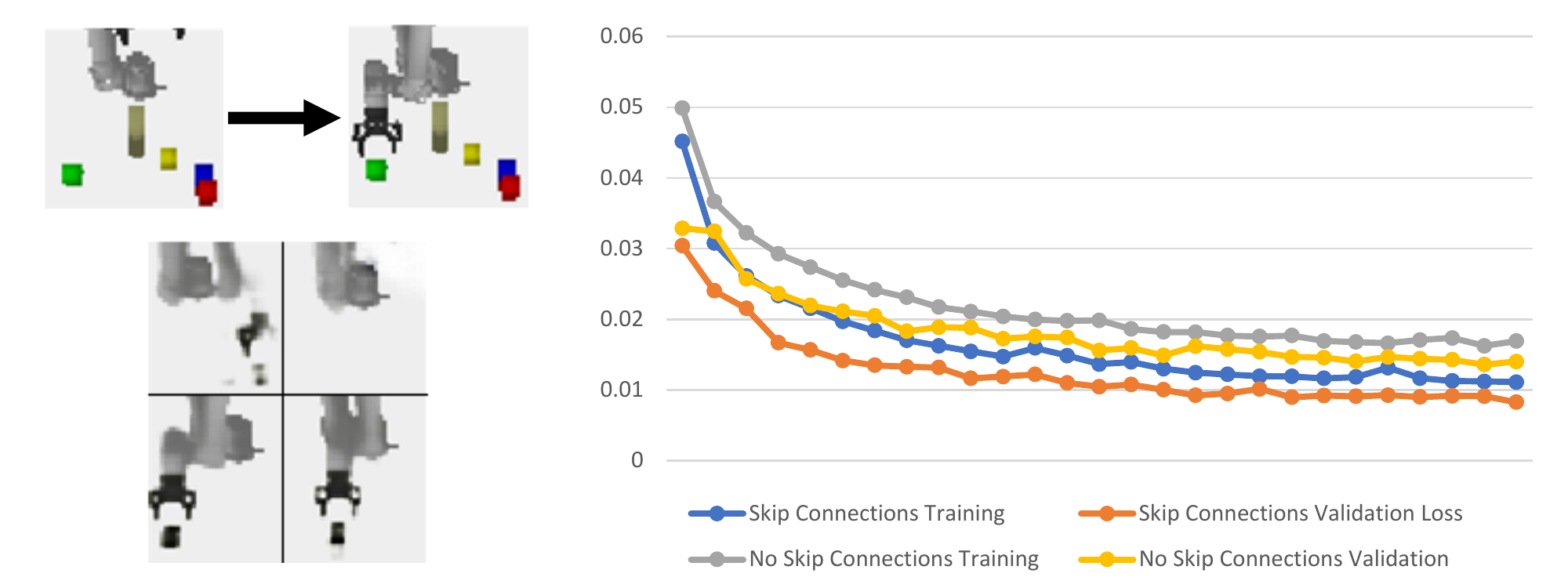}
	\caption{Without the skip connections, the model has a harder time learning what aspects of the scene are important and which will change over time. Left: example of images produced with no skip connections. Right: training and validation loss for the first 7,500 iterations.}
	\label{fig:no-skip-connections}
\end{figure*}

\paragraph{Skip Connections.}~\hspace*{-7pt}~We compare the error of our models both with and without skip connections between the encoder and the image decoder.
This version of the model was trained and evaluated on successful trials. Fig.~\ref{fig:no-skip-connections} shows an example of results without skip connections: the model was able to learn several possible positions, but image quality is subjectively far lower.

In addition, we compared absolute image and pose error for models trained with and without training data on both versions of the model. We see that without skip connections, we see an image-pixel error of $5.01 \times 10^{-4}$, with an average pose error of $9.61 \times 10^{-5}$. Total, weighted validation loss was $0.012$. By contrast, for the version with skip connections, we see image loss of $1.21 \times 10^{-4}$ and pose error of $5.51 \times 10^{-5}$.
In short, we can see that the skip connections make a difference when it comes to image reconstruction: images are both qualitatively higher quality, and have a quarter the pixel-wise error. In addition, end effector poses are roughly twice as accurate. 

Presumably, a large part of the advantage of skip connections is that the hidden state only needs to encode aspects of the image that are changing from one frame to the next. Fig.~\ref{fig:no-skip-connections}~(right) shows the effect this has on training: loss remained consistently higher for the version without skips.

\section{Conclusions}\label{sec:conc}

We described an approach for learning a predictive model that can be used to generate interpretable task plans for robot execution.
This model supports complex tasks, and requires only minimal labeling.
It can also be applied to many different domains with minimal adaptation.
We also provided an analysis of how to create and train models for this problem domain, and describe the validation of the final model architecture in a range of different domains.

Still, there are clear avenues for improvement in future work.
First, in this work we assume that low-level ``actor'' policies are provided.
This is sufficient for most manipulation and navigation tasks, but may not capture complex object interactions.
Second, we assume the existence of mid-level supervisory labels in this work to extract change-points.
In the future, we would prefer to detect change-points automatically using an approach such as that proposed by ~\cite{niekum2012learning}. Finally, we plan to expand this method into a full planning algorithm using predicted value and action priors that operates on sensor data.


\bibliographystyle{IEEEtran}
\bibliography{planners,task_description,taskmodels,drl,software,oac,ltl,representation,prediction,machine_learning}

\end{document}